%% file: example.tex
\documentclass{article}

\usepackage[preprint]{corl_2021} 

\usepackage{amsmath}
\usepackage{amssymb}
\usepackage{mathtools}
\usepackage{comment}
\usepackage{xcolor}
\usepackage[inline]{enumitem}

\DeclarePairedDelimiterX{\infdivx}[2]{(}{)}{%
  #1\;\delimsize\|\;#2%
}
\newcommand{\infdiv}{KL\infdivx}

\newlength\mywdE
\newcommand{\E}[3][]{\settowidth\mywdE{$\mathbb{E}$}
\mathop{\mathbb{E}}_{\vphantom{|^|}\mathmakebox[0.5\mywdE][l]{#2}}#1[#3#1]}
\newcommand{\pluseq}{\mathrel{+}=}

\usepackage{graphicx}
\usepackage[ruled,vlined,linesnumbered]{algorithm2e}
\title{Cycle-Consistent World Models for Domain Independent Latent Imagination}

\author{
  Sidney Bender \\
  TU Berlin (TUB) \\
  Berlin, Germany\\
  \texttt{s.bender@tu-berlin.de} \\
  \And
  Tim Joseph \\
  FZI Research Center for Information Technology \\
  Karlsruhe, Germany \\
  \texttt{joseph@fzi.de}
  \And
  J. Marius Zöllner \\
  Karlsruhe Institute of Technology (KIT)\\
  Karlsruhe, Germany\\
  \texttt{marius.zoellner@kit.edu} \\
}

\begin{document}
\maketitle


\begin{abstract}
End-to-end autonomous driving seeks to solve the perception, decision, and control problems
in an integrated way, which can be easier to generalize at scale and be more adapting to
new scenarios.
However, high costs and risks make it very hard to train autonomous cars in
the real world.
Simulations can therefore be a powerful tool to enable training.
Due to slightly different observations, agents trained and evaluated solely in simulation often perform well there but have difficulties in real-world environments.
To tackle this problem, we propose a novel model-based reinforcement learning approach called Cycle-consistent World Models.
Contrary to related approaches, our model can embed two modalities in a shared latent space and thereby learn from samples in one modality (e.g., simulated data) and be used for inference in different domain (e.g., real-world data).
Our experiments using different modalities in the CARLA simulator showed that this enables CCWM to outperform state-of-the-art domain adaptation approaches.
Furthermore, we show that CCWM can decode a given latent representation into semantically coherent observations in both modalities.
\end{abstract}

\input{content/introduction}
\input{content/prerequisites}

\input{content/approach}

\input{content/related_work}
\input{content/experiments}

\input{content/discussion}



\clearpage
\acknowledgments{The research leading to these results is funded by the German Federal Ministry for Economic Affairs and Climate Action" within the project “KI Delta Learning“ (Förderkennzeichen 19A19013L). The authors would like to thank the consortium for the successful cooperation.}

\bibliography{example, library}  

\end{document}

%% file: content/introduction.tex
\section{Introduction}

Many real-world problems, in our case autonomous driving, can be modeled as high-dimensional control problems. In recent years, there has been much research effort to solve such problems in an end-to-end fashion. While solutions based on imitation learning try to mimic the behavior of an expert, approaches based on reinforcement learning try to learn new behavior to maximize the expected future cumulative reward given at each step by a reward function. In a wide range of areas, reinforcement learning agents can achieve super-human performance \cite{dqn, alpha_star, mu_zero} and outperform imitation learning approaches~\cite{toromanoff2020end}. 

 However, for high-dimensional observation spaces many reinforcement learning algorithms that are considered state-of-the-art learn slowly or fail to solve the given task at all. Moreover, when the agent fails to achieve satisfactory performance for a given task, it is hard to analyze the agent for possible sources of failure. Model-based reinforcement learning promises to improve upon these aspects. Recent work has shown that model-based RL algorithms can be a magnitude more data-efficient on some problems \cite{pla_net, Kurutach2018Model-EnsembleOptimization, Janner2019WhenOptimization, Chua2018DeepModels, dreamer, dreamer_v2}. Additionally, since a predictive world model is learned, one can analyze the agent's perception of the world \cite{Chen2021InterpretableLearning}. 

Still, such agents are mostly trained in simulations \cite{Bellemare2013TheAgents, Brockman2016OpenAIGym, Tassa2018DeepMindSuite} since interaction with the real world can be costly (for example, the cost for a fleet of robots or the cost to label the data). Some situations should be encountered to learn, but must never be experienced outside of simulation (e.g., crashing an autonomous vehicle). While simulations allow generating many interactions, there can be a substantial mismatch between the observations generated by the simulator and the observations that the agent will perceive when deployed to the real world. Furthermore, observations from simulation and reality are mostly unaligned, i.e., there is no one-to-one correspondence between them. This mismatch is often called the domain gap \cite{Ganin2015UnsupervisedBackpropagation} between the real and simulated domain. When the domain gap is not taken into account, the behavior of an agent can become unpredictable as it may encounter observations in reality that have never been seen before in simulation.

One family of approaches to reduce this gap is based on the shared-latent space assumption \cite{unit}. The main idea is that the semantics of an observation are located in a latent space from which a simulated and an aligned real observation can be reconstructed. Approaches grounded on this assumption have recently been able to achieve impressive results in areas such as style transfers~\cite{huang2018multimodal} and imitation learning~\cite{Bewley2018LearningLabels}.

Inspired by this, we propose adopting the idea of a shared latent space to model-based reinforcement learning by constructing a sequential shared-latent variable model. Our main idea is to create a model that allows to plan via latent imagination independently of the observation domain. The model is trained to project observation sequences from either domain into a shared latent space and to predict the future development in this latent space. By repeatedly rolling out the model one can then plan or train a policy based on low-dimensional state trajectories. 

Our contributions can be summarized as follows: 
\begin{enumerate*}
    \item We present a novel cycle-consistent world model (CCWM) that can embed two similar partially observable Markov decision processes that primarily differ in their observation modality into a shared latent space without the need for aligned data. 
    \item We show that observation trajectories of one domain can be encoded into a latent space from which CCWM can decode an aligned trajectory in the other domain. This can be used as a mechanism to make the agent interpretable.
    \item We test our model in a toy environment and train a policy via latent imagination first and then evaluate and show that it is also able to learn a shared latent representation for observations from a more complex environment based on the CARLA simulator.

\end{enumerate*}

%% file: content/prerequisites.tex
\section{Preliminaries}

\textbf{Sequential Latent Variable Models}
In contrast to model-free reinforcement learning (RL), model-based RL explicitly learns an approximate transition model of the environment to predict the next observation $x_{t+1}$ from the current observation $x_{t}$ and the chosen action $a_{t}$ \cite{Sutton2018ReinforcementIntroduction}. The model is used to rollout imagined trajectories $x_{t+1},a_{t+1},x_{t+2},a_{t+2}, ... $ which can be either used to find the best future actions or to train a policy without the need to interact with the real environment. A problem with such a model is that rollouts become computationally expensive for high-dimensional observation spaces. For this reason, many recent model-based RL algorithms make use of sequential latent variable models. Instead of learning a transition function in observation space $X \subseteq \mathbb{R}^{d_X}$, observations are first projected into a lower-dimensional latent space $S \subseteq \mathbb{R}^{d_S}$ with $d_S \ll d_X$ . Then a latent transition function can be used to rollout trajectories of latent states $s_{t+1},a_{t+1},s_{t+2},a_{t+2}, ... $ computationally efficient \cite{Hafner2018LearningPixels, Hafner2020DreamImagination}. Since naive learning of latent variable models is intractable, a prevailing way to train such models is by variational inference \cite{Kingma2014Auto-EncodingBayes}. The resulting model consists of the following components:

\begin{itemize}
    \item Dynamics models: prior $p_{\theta}(s_t | s_{t-1}, a_{t-1})$ and posterior $q_{\theta}(s_t | s_{t-1}, a_{t-1}, x_t)$
    \item Observation model: $p_{\theta}(x_t | s_t)$
\end{itemize}

Furthermore, at each time step the resulting loss function encourages the ability to reconstruct observations from the latent 
states while at the same time enforcing to be able to predict the future states from past observations. This loss function is also known as the negative of the evidence lower bound (ELBO):

\begin{equation}\label{eq:sequential-variable-model-loss}
    L_t = 
    \underbrace{-\E{q_{\theta}(s_t | x_{\leq t}, a_{\leq t})}{p_{\theta}(x_t|s_t)}}_{\text{reconstruction loss} \ L_{\text{recon}}}
    \quad 
    +
    \quad 
    \underbrace{
    \E{q_{\theta}(s_{t-1} | x_{\leq {t-1}}, a_{\leq {t-1}})}{\infdiv{q_{\theta}(s_t | s_{t-1}, a_{t-1}, x_t)}{p_{\theta}(s_t | s_{t-1}, a_{t-1})}}
}_{\text{regularization loss} \ L_{\text{reg}}}
\end{equation}

\textbf{Shared Latent Space Models}
We want to enable our model to jointly embed unaligned observation from two different modalities of the same partially observable Markov decision process into the same latent space.
Let $X_A$ and $X_B$ be two observation domains (e.g., image domains with one containing RGB images and the other one containing semantically segmented images).
In aligned domain translation, we are given samples $(x_B, x_B)$ drawn from a joint distribution $P_{X_A,X_B} (x_A, x_B)$.
In unaligned domain translation, we are given samples drawn from the marginal distributions $P_{X_A} (x_A)$ and $P_{X_B} (x_B)$.
Since an infinite set of possible joint distributions can yield the given marginal distributions, it is impossible to learn the actual joint distribution from samples of the marginals without additional assumptions.

A common assumption is the shared-latent space assumption \cite{Liu2017UnsupervisedNetworks, Liu2016CoupledNetworks}.
It postulates that for any given pair of samples $(x_A$, $x_B) \sim P_{X_A, X_B}(x_A, x_B) $ there exists a shared latent code $s$ in a shared-latent space such that both samples can be generated from this code, and that this code can be computed from any of the two samples.
In other words, we assume that there exists a function with $s = E^{A \to S}(x_A)$ that maps from domain $X_A$ to a latent space $S$ and a function with $x_A = G^{S \to A}(s)$ that maps back to the observation domain. Similarly, the functions $s = E^{B \to S}(x_B)$ and $x_B = G^{S \to B}$ must exist and map to/from to the same latent state .

Directly from these assumptions follows that observations of domain $A$ can be translated to domain $B$ via encoding and decoding and the same must hold for the opposite direction:
\begin{equation} \label{eq:shared-latent-space:translation}
    \begin{aligned} 
        G^{S \to B}(E^{A \to S}(x_A)) \in X_B \\
        G^{S \to A}(E^{B \to S}(x_A)) \in X_A
    \end{aligned}
\end{equation}

Another implication of the shared latent space assumption is that observations from one domain can be translated the other one and back to the original domain (\textbf{cycle-consistency} \cite{Zhu2017UnpairedNetworks}):
\begin{equation} \label{eq:shared-latent-space:cyclic}
    \begin{aligned}
        E^{A \to S}(x_a) = E^{B \to S}(G^{S \to B}(E^{A \to S}(x_A))) \\
        E^{B \to S}(x_b) = E^{A \to S}(G^{S \to A}(E^{B \to S}(x_B)))
    \end{aligned}
\end{equation}

The fundamental idea is that by enforcing both of them on semantically similar input domains, the model embeds semantically similar samples close to each other in the same latent space.

%% file: content/approach.tex
\section{Cycle-consistent World Models}


In this section, we present our cycle-consistent world model (CCWM). Considering the structure of sequential latent variable models and the constraints resulting from the shared latent space assumption, we show how both can be integrated into a single unified model. In the following, we explain the model architecture and the associated loss terms.

\begin{figure}[h]
\includegraphics[width=\textwidth]{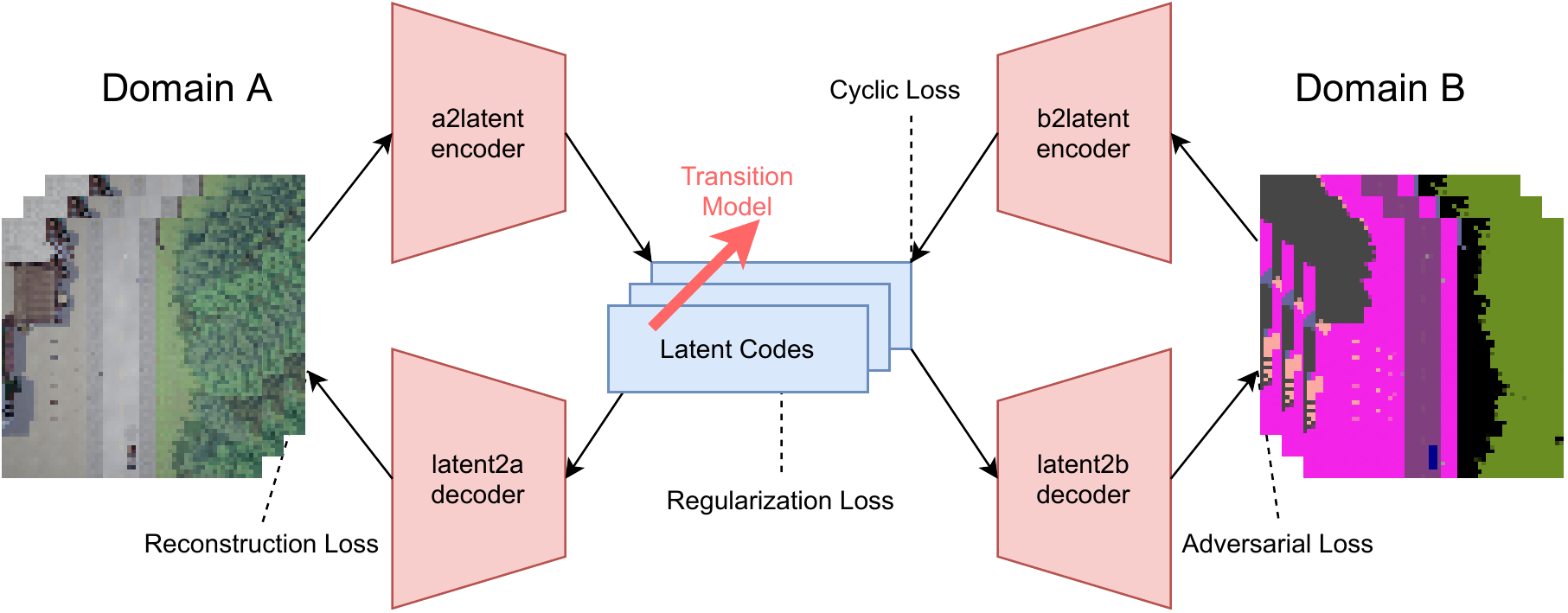}
\label{fig:ccwm}
\caption{
Cycle-consistent world model.
In the pictured situation, a sequence of top camera images is used a the input.
The images are encoded frame-wise into latent states and forward predicted by the transition model.
From these latent codes, reconstructed top camera images and images translated to semantic top camera images are calculated.
From the translated images, cyclic latent codes are calculated.
Finally, the four losses can be calculated, which enforce equations ~(\ref{eq:shared-latent-space:translation}) and (\ref{eq:shared-latent-space:cyclic}).
}
\end{figure}

\textbf{Architecture}
Since our model is a sequential latent variable model, it includes all the components that have been presented in section 2, namely the prior transition model $p_{\theta}(s_t | s_{t-1}, a_{t-1})$, the posterior transition model $q_{\theta}(s_t | s_{t-1}, a_{t-1}, h_t)$ and an observation model $p^A_{\theta}(x_t | s_t)$ with $Dec_A(s_t)=\operatorname{mode}(p^A_{\theta}(x_t | s_t))$. Additionally, we define a feature extractor with $h_t = Enc_A(x_t)$ and a reward model $p^A_{\theta}(r_t | s_t)$.
So far, this model can be used as the basis of an RL-agent that acts on a single domain by first building up the current latent representation $s_t$ using the feature extractor and posterior and then rolling out future trajectories $s_{t+1}, s_{t+2}, ...$ with their associated rewards with the prior dynamics and the reward model. 
To project to and from another domain $X_B$ into the same latent space $S$ we add another feature extractor $Enc_B(x_t)$ and observation model $p^B_{\theta}(x_t | s_t)$ with $Dec_B(s_t)=\operatorname{mode}(p^B_{\theta}(x_t | s_t))$. Both are similar to their domain $X_A$ counterparts but do not share any weights. The prior dynamics model is shared since it does not depend on observation. In contrast, we need another posterior dynamics model for domain $B$, but since we let it share weights with its domain $A$ counterpart, we effectively only have a single posterior dynamics model. Additionally, we add a reward model $p_{\theta}(r_t|s_t)$ that also is shared between both domains so that latent trajectories can be rolled out independently of the observation domain. A major advantage of this approach is that we can train a policy with our model without regard to the observation domains. 

Finally, for training only, we need two discriminators $Dis^A_\phi$ and $Dis^B_\phi$ to distinguish between real and generated samples for each domain. It is important to note that the discriminators have a separate set of parameters $\phi$.

\textbf{Losses}
Given a sequence of actions and observations $\{a_t, x_t\}_{t=k}^{k+H} \sim D_A$ from a dataset $D_{A}$ collected in a single domain $X_A$, we first roll out the sequential latent variable model using the posterior to receive an estimate for the posterior distribution $q(s_t | s_{t-1}, a_{t-1}, x_t)$ and the prior distribution $q(s_t | s_{t-1}, a_{t-1}, x_t)$ for each time step. We can then calculate the following losses:
$L_{\text{recon}}$ is the reconstruction loss of the sequential latent variable model and $L_{\text{reg}}(q, p) = \infdiv{q}{p}$ is the regularization loss that enforces predictability of futures state as shown in equation \ref{eq:sequential-variable-model-loss}.
$L_{\text{adv}}(x) = Dis_B(x)$ is an adversarial loss that penalizes translations from domain $X_A$ to $X_B$ via $S$ that are outside of domain $X_B$ to enforce equation \ref{eq:shared-latent-space:translation} of the shared latent space assumption. Here, $Dis_B$ is a PatchGAN~\cite{isola2017image} based discriminator that is trained alongside our model to differentiate between real and generated observations.
The cycle loss $L_{\text{cyc}}(q, p) = \infdiv{q}{p}$ is derived from the cycle constraints of equation \ref{eq:shared-latent-space:cyclic} and calculates the KL-divergence between the posterior state distributions conditioned on observations and states from domain $A$ and conditioned on observations and states that have been translated to domain $B$, i.e. $x_t \to s_t \to x^\text{trans}_t \to s_t^\text{cyc}$ (see algorithm \ref{algorithm}; line 7, 8 and 12). To calculate the cyclic loss it is necessary to roll out a second set of state trajectories using the cyclic encoding $h_t^{\text{cyc}}$ and the cyclic state $s_t^\text{cyc}$.

For sequences of domain $B$, we train with the same loss functions, but with every occurrence of $A$ and $B$ interchanged. This is also shown in algorithm \ref{algorithm}line 26 and line 28.

\begin{algorithm}[h]
\SetAlgoLined
\textbf{Input:}
Replay Buffers $D_A$ and $D_B$,
Encoders \textcolor{blue}{$Enc_a$} and \textcolor{red}{$Enc_b$},
Decoders \textcolor{blue}{$Dec_a$} and \textcolor{red}{$Dec_b$},
Model parameters $\Theta$, 
Discriminator parameters $\Phi$\; 
\SetKwProg{Fn}{Function}{:}{}
  \Fn{$L_{gen}$($Enc_1$, $Dec_1$, $Enc_2$, $Dec_2$, $x_{1:T}$)}{
  \ForEach {$t \in \mathcal T$}{
    $h_t \leftarrow Enc_1(x_t)$\;
    $s_t' \sim q(s_t|s_{t-1}',h_t)$\;
    $x^{\text{recon}}_{t} \leftarrow Dec_1(s_t')$\;
    $x^{\text{trans}}_{t} \leftarrow Dec_2(s_t')$\;
    $h^{\text{cyc}}_{t} \leftarrow Enc_2(x^{\text{trans}}_{t})$\;
    $s^{\text{cyc}}_t \sim$
    $q(s^{\text{cyc}}_t|s^{\text{cyc}}_{t-1},h^{\text{cyc}}_t)$\;
    $L_{\text{ret}} \pluseq L_{\text{recon}}(x_t,x^{\text{recon}}_t)$\;
    $L_{\text{ret}} \pluseq L_{\text{adv}}(x^{\text{trans}}_t)$\;
    $L_{\text{ret}} \pluseq L_{\text{reg}}(q(s_t|s_{t-1}',h_t), p(s_t|s_{t-1}'))$\;
    $L_{\text{ret}} \pluseq L_{\text{cyc}}(q(s_t|s_{t-1}',h_t),$ $q(s_{t}|s^{\text{cyc}}_{t-1}, h^{\text{cyc}}_{t}))$\;
  }
  \KwRet $L_{\text{ret}}$\;
 }
\SetKwProg{Def}{Function}{:}{}
  \Def{$L_{\text{dis}}$($Enc_1$, $Dec_2$, $x^1$, $x^2$)}{
  \ForEach {$t \in \mathcal T$}{
    $h_t \leftarrow Enc_1(x^1_t)$\;
    $s_t \sim q(s_t|s_{t-1},h_t)$\;
    $x^{\text{trans}}_{t} \leftarrow Dec_2(s_t)$\;
    $L_{\text{ret}} \pluseq L_{\text{adv}}(x^2_t) + (1 - L_{\text{adv}}(x^{\text{trans}}_t)) $\;
    }
    \KwRet $L_{\text{ret}}$\;
 }
 \While{not converged}{
  Draw sequence of $x_{a,1:T} \sim D_A$\;
  Draw sequence of $x_{b,1:T} \sim D_B$\;
  $L_{gen} = L_{gen}(
  \textcolor{blue}{Enc_a}, \textcolor{blue}{Dec_a}, \textcolor{red}{Enc_b}, \textcolor{red}{Dec_b}, x_{a,1:T}
  ) + L_{gen}(
  \textcolor{red}{Enc_b}, \textcolor{red}{Dec_b}, \textcolor{blue}{Enc_a}, \textcolor{blue}{Dec_a}, x_{b,1:T}
  )$\;
  Update Model parameters $\Theta \leftarrow \Theta + \Delta L_{gen}$\;
  $L_{\text{dis}} = L_{\text{dis}}(\textcolor{blue}{Enc_a}, \textcolor{red}{Dec_b}, x_{a,1:T}, x_{b,1:T}) + L_{\text{dis}}(\textcolor{red}{Enc_b}, \textcolor{blue}{Dec_a}, x_{b,1:T}, x_{a,1:T})$\;
  Update Discriminator parameters $\Phi \leftarrow \Phi + \Delta L_{dis}$\;
 }
 \caption{Training Routine of the CCWM}
 \label{algorithm}
\end{algorithm}

%% file: content/related_work.tex
\section{Related Work}


\textbf{Control with latent dynamics}
World Models~\cite{ha2018world} learn latent dynamics in a two-stage process to evolve linear controllers in imagination.
PlaNet~\cite{Hafner2018LearningPixels} learns them jointly and solves visual locomotion tasks by latent online planning. Furthermore, Dreamer \cite{Hafner2020DreamImagination, dreamer_v2} extends PlaNet by replacing the online planner with a learned policy that is trained by back-propagating gradients through the transition function of the world model. 
MuZero~\cite{mu_zero} learns task-specific reward and value models to solve challenging tasks but requires large amounts of experience.
While all these approaches achieve impressive results, they are limited to their training domain and have no inherent way to adapt to another domain.

\textbf{Domain Randomization}
\citeauthor{James2019Sim-to-realNetworks}~\cite{James2019Sim-to-realNetworks} introduce a novel approach to cross the visual reality gap, called Randomized-to-Canonical Adaptation Networks (RCANs), that uses no real-world data.
RCAN learns to translate randomized rendered images into their equivalent non-randomized, canonical versions.
In turn, this allows for real images to be translated into canonical simulated images.
\citeauthor{random_convolutions}~\cite{random_convolutions} showed that random convolutions (RC) as data augmentation could greatly improve the robustness of neural networks.
Random convolutions are approximately shape-preserving and may distort local textures.
RC outperformed related approaches like~\cite{volpi2018generalizing, qiao2020learning, wang2019learning} by a wide margin and is thereby considered state-of-the-art by us.

\textbf{Unsupervised Domain Adaptation}
The original Cycle-GAN~\cite{Zhu2017UnpairedNetworks} learn to translate images from one domain to another by including a a cycle loss and an adversarial loss into training.
\citeauthor{Liu2017UnsupervisedNetworks}~\cite{Liu2017UnsupervisedNetworks} extend this idea with weight sharing of the inner layers and a normalization loss in the latent state, which enables it to embed images of semantically similar domains into the same latent space.
\textit{Learning to drive}~\cite{ltd} uses this idea to train an imitation learning agent in simulation and successfully drive in reality.
In \textit{RL-Cycle-GAN}~\cite{rl_cycle_gan}, a Cycle-GAN with an RL scene consistency loss is used, and the authors show that even without the RL scene consistency loss, RCAN~\cite{sim2sim} was outperformed by a wide margin.
RL-Cycle-GAN is state-of-the-art for unsupervised domain adaptation to the best of our knowledge.

%% file: content/experiments.tex
\section{Experiments}

First, we will demonstrate our model in a small toy environment. Then we will show its potential in a more realistic setting related to autonomous driving based on the CARLA simulator ~\cite{dosovitskiy2017carla}.

\textbf{Implementation}
Our prior and posterior transition models are implemented as recurrent state-space models (RSSM)~\cite{Hafner2018LearningPixels}.
In the RSSM, we exchanged the GRU~\cite{} with a convolutional GRU~\cite{siam2017convolutional}.
A challenge of integrating the ideas of a world model and a shared latent space assumption is that it is easier to enforce a shared latent space on a large three-dimensional tensor-shaped latent space.
In contrast, most world models use a low-dimensional vector latent space.
A bigger latent space makes it easier to embed and align both modalities, but it leads to a less informative self-supervised encoding for the downstream heads, such as the reward model.
As we show in our ablation study choosing the right height and width of the latent space is crucial for successful learning.

\begin{figure}[h]
\centering
\includegraphics[width=1.0\textwidth]{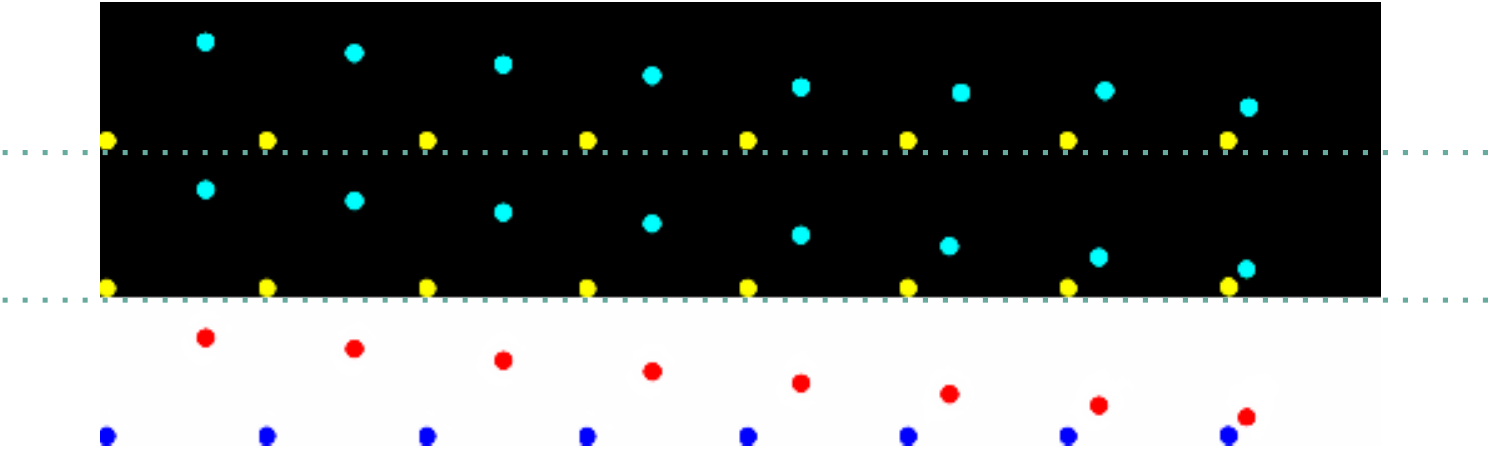}
\label{fig:dreamer}
\caption{
Qualitative results on ArtificialV0.
The top row shows the observations recorded from the environment if one observation is given to the model and the policy is rolled out.
It shows that the model can learn the optimal policy (bringing the red/turquoise dot towards the blue/yellow dot on a straight line) only with downstream information from the original modality but also works in the reversed modality.
The second row is the prediction of our CCWM back into the domain from that the agent retrieved the initial observation.
The last row is the cross-modality prediction.
}
\label{fig:artificial_v0_qualitative}
\end{figure}

\textbf{Proof of Concept}
Reinforcement learning environments are often very complex, so that evaluation and model analysis can become hard for complex models such as ours.
Additionally, domain adaptation complicates evaluation even more.
For this reason, we first construct a toy environment that we call ArtificialV0 to show that our idea is working in principle.
ArtificialV0 is constructed as follows:
A state of ArtificialV0 is the position of a red and a blue dot.
Its state space is a box $[-1,1] \times [-1,1]$.
As observations, we use images of the red and the blue dot on a white background.
The goal is to move the red dot towards the blue dot.
The actions are steps by the red dot with an action space of $[-0.2,0.2] \times [-0.2,0.2]$.
The negative euclidean distance between the blue and the red dot is used as a reward.
An episode terminates as soon as the absolute Euclidean distance is smaller than $0.1$.
The other modality is constructed the same, but the observation images are inverted.
Advantages of ArtificialV0 are that the actions and observations are easy to interpret and the optimal policy as a reference benchmark is easy to implement.
The optimal policy brings the red dot on a straight line towards the blue dot and achieves an average return of $-2.97$.
We find that CCWM achieves a similar average return after 30K environment steps in an online setting in both modalities, despite us only giving it access to a small offline dataset of 5000 disjunct observations from the reversed modality without downstream information.
In figure~\ref{fig:artificial_v0_qualitative}, one can see that a trajectory can be started in the inversed modality and successfully continued in both modalities.
This indicates that the model is capable of embedding both modalities into a shared latent space.

\textbf{Experiment Setup}
To show the potential of our approach in a more realistic environment, we also evaluate our model in the CARLA simulator.
We choose to use images from a semantic camera as the first modality and images from an RGB camera as the second modality.
Both look down onto the cars from a birds-eye-view point.

For an even more realistic setting, one could replace the top view RGB camera with an RGB surround camera in a real car and the schematic top view with an RGB surround-view camera from in simulation. However, since we do not have access to a real car with such sensors and we are restricted in computational resources, we simplified the problem for now.
Arguably, the visual difference between the RGB camera from the simulation and the real world RGB camera is smaller than the visual difference between the RGB camera in the simulation and the schematic view of the simulation, so there is reason to believe that a transfer from the RGB camera of the simulation to the RGB camera of the real world would work as well.

\begin{table}[h]
\centering
\begin{tabular}{ | c | c c | c |}
\hline
 Approach           & Reward RSE    & Reward RSE cross-modality          & PSNR      \\
 \hline
 Single Modality    & 0.25          & 3.86                          & 10.21     \\ 
 RC                 & 0.31          & 0.49                          & 11.39     \\  
 CycleGAN           & 0.28          & 0.57                          & 12.28     \\  
 \hline
 Ours               & 0.23          & \textbf{0.48}                 & \textbf{13.91}     \\
 \hline
\end{tabular}
\newline
\caption{
Comparison with the state-of-the-art.
We measured the quality of the reward prediction with the relative squared error against predicting the mean reward to show that something better than predicting the mean is learned.
Furthermore, we determined how well the different models can predict the next states based on the peak signal-to-noise ratio (PSNR) between the real future observations and the predicted observations.
We can see that all domain adaptation methods can transfer the reward predictions while only using one modality.
Our CCWM achieved the best reward transfer and the best video prediction.
It is worth mentioning that the cross-modality reward predictions with only one modality and with RC were unstable, varying strongly over time steps depending on the initialization.
Since single modality and RC are fast, while Cycle-GAN and CCWM are slow, we show the results after training approximately 24 hours on an NVIDIA GTX1080TI to keep the comparison fair.
}
\label{table:state_of_the_art}
\end{table}

\begin{table}[b]
\centering
\begin{tabular}{ | c | c c | c |}
\hline
 Latent Space Size  & Reward RSE    & Reward RSE cross-modality     & PSNR      \\
 \hline
 $1 \times 1$       & 0.92          & 1.18                          & 13.00     \\ 
 $2 \times 2$       & 0.95          & 1.10                          & 13.80     \\  
 $4 \times 4$       & 0.57          & 0.57                          & 13.81     \\  
 $8 \times 8$       & 0.23          & \textbf{0.48}                 & \textbf{13.91}     \\
 \hline
\end{tabular}
\newline
\caption{
Ablation study on the size of the latent space.
The models are identical except that the convolutional GRU is used at different downsampling scales of the network.
We can see that latent spaces smaller than $4 \times 4$ are having trouble minimizing all objectives at once, and the reward RSE is not falling significantly below simply predicting the mean.
}
\label{table:ablation_study}
\end{table}

\textbf{Comparsion with the state-of-the-art}
To show that the constructed domain gap is not trivial and our model is outperforming current domain adaptation methods, we compare our model with 1) no adaptation to the other modality at all, 2) the random convolutions (RC)~\cite{random_convolutions} approach, which we regard as being state of the art in domain randomization, and 3) the RL-CycleGan~\cite{rl_cycle_gan}, which we consider to be the start of the art in unsupervised domain adaptation.
All models are reimplemented and integrated into our codebase. They are apart from their core idea as similar as possible regarding network structure, network size, and other hyperparameters.
The performance of a world model rises and falls with two factors:
1) How well the model can predict the current reward based on the current state and
2) how accurate the prediction of the next states is.
We recorded three disjunct offline datasets with the CARLA roaming agent (an agent controlled by the CARLA traffic manager).
The first contains trajectories of observations of the semantic view with downstream information. The second contains trajectories of observations of the RGB camera without downstream information. The third contains aligned semantic and RGB camera trajectories and downstream information.
The first and the second dataset are used for training the model, and the third is used for evaluation.
The model without any domain adaptation is trained on the first dataset in the regular dreamer style for the model training.
The RC model is trained on the first dataset with randomized inputs.
The RL-Cycle-GAN model is trained by first learning a Cycle-GAN-based translation from the first modality to the second modality. Then the model is trained on the translated observations of the first dataset.
CCWM is trained as described in the previous section on the first and the second dataset.

\begin{figure}[h]
\includegraphics[width=\textwidth]{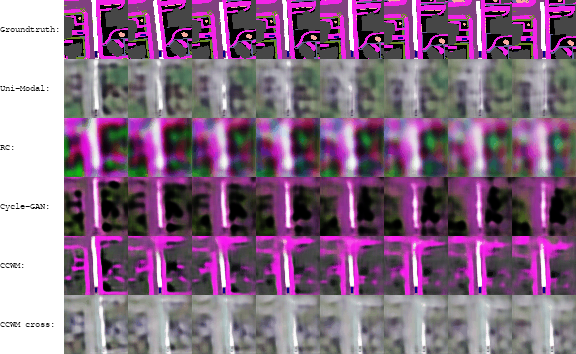}
\label{fig:dreamer}
\caption{
Qualitative Results on Carla.
The first row shows the ground truth of the semantic top camera sampled from dataset 3, and the second row the baseline of what would happen if the dreamer was trained in one modality and rolls out the other modality now.
Row 3 and 4 show the state-of-the-art comparison with random convolutions and a preprocessing input with a Cycle-GAN.
Both were also only trained on with the RGB top camera.
The 5th and the 6th row shows our model rolled out aligned in both modalities.
The previous 19 frames and the first frame of the ground truth are fed into the model for all models, and then the model is rolled out for fifteen time steps (every second is shown).
}
\label{fig:carla_qualitative}
\end{figure}

\textbf{Results}
All models are evaluated on the third dataset in the following ways:
First, we qualitatively analyze the predictive power for the next states of the model.
We warm up the model by feeding it some observations and then predict the next observations of the target domain, as shown in figure~\ref{fig:carla_qualitative}.
A general advantage of CCWM noteworthy to mention is that it can predict into both modalities simultaneously since both have a shared latent representation, which might be practical for error search.
Besides the qualitative analysis of the state predictions based on the predicted observations, we also compare the predictions quantitatively by calculating the PSNR between the predicted and the real trajectory, as seen in the table~\ref{table:state_of_the_art}.
Furthermore, we compare the reward prediction in the target domain where no downstream information was available.
Both in qualitative and quantitative comparison, one can see that our model outperforms the other approaches.

\textbf{Analysis}
The advantage of our approach over RC is that RC generalizes random distortions of the input image that RC can emulate with a random convolution layer, which might include the semantic segmentation mask, but will also include many other distributions, making it less directed despite its simplicity.
Pre-translating with Cycle-GAN follows a more directed approach but is not able to train the whole network end-to-end.
Furthermore, it first encodes a training image, then decodes it to a different domain, and then encodes it again to derive downstream information and predict future states.
This is a longer path than encoding it only once like CCWM and leaves room for well-known problems with adversarial nets like artifacts in the image, hindering training progress.

\textbf{Ablation Study}
Although probabilistic graphic models and reinforcement learning approaches are generally susceptible to hyperparameters, the size of the latent space has shown to be especially significant.
As shown in table~,\ref{table:ablation_study} a 1x1 latent space like it is common in many model-based RL approaches performs poorly, while bigger latent spaces provide much better performance.
Our explanation for this is twofold.
Firstly, related approaches such as UNIT \cite{Liu2017UnsupervisedNetworks} cannot translate images well with a tiny latent space and instead use huge latent spaces.
Secondly, in autonomous driving, it might not be beneficial to compress the whole complicated scene with multiple cars that all have their own location, direction, speed, etc. into one vector, but give the network inductive bias to represent each of them in a single vector and calculate the dynamics through the convolutional GRU with its suiting local inductive bias.
Another important consideration is the weights for the different losses, which need to be carefully chosen.
The reward loss tends to get stuck around the mean since its signal is relatively weak, so it should be chosen relatively high.
The KL based losses in the latent space can get very high and destroy the whole model with a single step.
On the other hand, a high normalization loss leads to bad predictive capabilities, and a high cyclic loss leads to a bad alignment of the modalities.

%% file: content/discussion.tex
\section{Conclusion}

In this work, we introduced cycle-consistent world models, a world model for model-based reinforcement learning that is capable of embedding two modalities into the same latent space.
We developed a procedure to train our model and showed its performance in a small toy environment and a more complex environment based on the CARLA simulator.
Furthermore, we compared it in an offline setting with two state-of-the-art approaches in domain adaptation, namely RC and RL-Cycle-GAN.
We outperformed RC by being more directed and Cycle-GAN by training end-to-end without the necessity to encode twice.
For the future we plan to extend our model by training a full model-based RL agent that is able to learn to control a vehicle in simulation and generalize to reality given only offline data from reality without any reward information.